\documentclass[runningheads]{llncs}
\usepackage{graphicx}

\usepackage{tikz}
\usepackage{comment}
\usepackage{amsmath,amssymb} 
\usepackage{color}
\usepackage{subcaption}
\usepackage{multirow}
\RequirePackage{cite} 
\usepackage{tabularx}
\usepackage{adjustbox,lipsum}
\usepackage{booktabs}

\usepackage[accsupp]{axessibility}  

\begin{document}
\pagestyle{headings}
\mainmatter
\def\ECCVSubNumber{100}  

\title{Identity-Aware Hand Mesh Estimation and Personalization from RGB Images} 

\titlerunning{Identity-Aware Hand Mesh Estimation and Personalization}
%
\author{Deying Kong\inst{1}\and
Linguang Zhang\inst{2} \and
Liangjian Chen\inst{2} \and
Haoyu Ma\inst{1}\and
Xiangyi Yan\inst{1}\and
Shanlin Sun\inst{1}\and
Xingwei Liu\inst{1}\and
Kun Han\inst{1} \and
Xiaohui Xie\inst{1}
}
\authorrunning{D. Kong et al.}
%
\institute{University of California - Irvine, Irvine CA 92697, USA \\
\email{\{deyingk, haoyum3, xiangyy4, shanlins, xingweil, khan7, xhx\}@uci.edu}\\ \and
Reality Labs at Meta, USA\\
\email{\{linguang,clj\}@fb.com}}
\maketitle

\begin{abstract}
Reconstructing 3D hand meshes from monocular RGB images has attracted increasing amount of attention due to its enormous potential applications in the field of AR/VR.
Most state-of-the-art methods attempt to tackle this task in an anonymous manner. Specifically, the identity of the subject is ignored even though it is practically available in real applications where the user is unchanged in a continuous recording session.
In this paper, we propose an \emph{identity-aware} hand mesh estimation model, which can incorporate the identity information represented by the intrinsic shape parameters of the subject.
We demonstrate the importance of the identity information by comparing the proposed \emph{identity-aware} model to a baseline which treats subject anonymously. 
Furthermore, to handle the use case where the test subject is \emph{unseen}, we propose a novel personalization pipeline to calibrate the intrinsic shape parameters using only a few unlabeled RGB images of the subject.
Experiments on two large scale public datasets validate the  state-of-the-art performance of our proposed method.


\end{abstract}

\section{Introduction}
Hand pose estimation has been one of the most popular computer vision problems because of its critical role in many applications, including hand gesture recognition, virtual and augmented reality, sign language translation and human-computer interaction~\cite{beddiar2020vision}.
%
With recent advances in deep learning techniques~\cite{he2016deep, simonyan2014very, newell2016stacked} and development of large hand pose datasets~\cite{simon2017hand, yu2020humbi, zimmermann2019freihand, zb2017hand}, 2D hand pose estimation has been extensively investigated and deployed in real-time applications with compelling results~\cite{cao2019openpose, kong2019adaptive,  simon2017hand}.
However, 3D hand pose estimation still remains a challenging problem due to the diversity of hand shapes, occlusion and depth ambiguity when monocular RGB image is used. 

Current state-of-the-art methods for 3D hand reconstruction from RGB images either try to directly regress 3D vertices of the hand mesh~\cite{ge20193d, kulon2020weakly, chen2021camera, lin2021end, lin2021mesh}, or utilize the parametric MANO model~\cite{romero2017embodied} by regressing the low-dimensional parameters~\cite{boukhayma20193d, zhang2019end, baek2019pushing, hasson2019learning, yang2020bihand}.
While these methods could generalize reasonably across different subjects, nearly all of them estimate the 3D hand pose in an anonymous manner. 
The identity information of the subject, which is practically available in real applications, is typically ignored in these methods. In many real-world use cases, such as virtual and augmented reality, the device is often personal and the user is typically identifiable. 

We ask the question, can 3D hand reconstruction from RGB images be further improved with the help of identity information? 
If so, how should we calibrate the personalized hand model for \emph{unseen} subjects during the test phase, using only RGB images?
In depth image based  hand tracking systems, the hand model personalization has been well studied and its benefits on improving hand tracking performance has been demonstrated~\cite{tan2016fits, tkach2017online}.
However, using only RGB images to perform personalization is underexplored. 

To close this gap and answer the above question,  we investigate the problem of hand model personalization from RGB images and design a simple yet effective network to incorporate the identity information. 
Specifically, we propose an \emph{identity-aware} hand mesh estimation model, which can take in the personalized hand model along with the input RGB image. 
Motivated by MANO~\cite{romero2017embodied}, we choose to use MANO shape parameters to represent the hand model.
To enable a fair comparison, we then construct a strong baseline by adapting our proposed identity-aware network slightly. Instead of being given the groundtruth hand shape parameters, the baseline regresses the shape parameters directly from the input image via a multi-layer perceptron.
We show through experiments that with ground truth shape parameters, more accurate 3D hand reconstruction can be obtained. 
Lastly, we propose a novel personalization method which can calibrate the hand model for \emph{unseen} subjects, using only unannotated RGB images. The calibrated hand model can then be utilized in our identity-aware network.
Our main contributions are summarized as follows:

\begin{itemize}
  \item Our work is the first to systematically investigate the problem of hand mesh personalization from RGB images and demonstrate its benefits to hand mesh and keypoints reconstruction.
  
  \item For unknown subjects that are \emph{not seen} in training, we develop a novel hand model personalization method that is capable of calibrating the hand model using a few unannotated images of the same subject.

  \item We demonstrate that our method outperforms existing methods on two large-scale public datasets, showing the benefit of utilizing the identity information, which is an underexplored topic in the field.
  
  \item We design a simple but competitive baseline that features the same optimization augmented inference step and further validate the effectiveness of leveraging the identity information.
\end{itemize}

\section{Related Work}

There are many research works on human/hand pose estimation~\cite{zhebestofboth, zhecross, zheGPA, lin2021end, lin2021mesh, moon2020i2l, moon2018v2v, ma2021transfusion, haoyu22ppt}, including well-developed 2D hand pose estimation algorithms~\cite{wang2018mask, kong2019adaptive, cao2019openpose, simon2017hand, kong2020rotation, chen2020nonparametric,  kong2020sia} and fast developing 3D hand pose estimation algorithms~\cite{athitsos2003estimating, zimmermann2017learning, Ge_2018_ECCV, cai2018weakly, spurr2020weakly}. In this section, we will mainly discuss literature on 3D hand mesh reconstruction. 

{\bf Model-based methods.}
The popular model-based method usually rely on the MANO model~\cite{romero2017embodied}, developed from the SMPL human model~\cite{loper2015smpl}.
As a parameterized model, the MANO model factorizes the hand mesh into shape and pose parameters, by utilizing principal component analysis.
Massive literature has tried to predict the MANO parameters in order to reconstruct the hand mesh. Boukhayma~\emph{et al}~\cite{boukhayma20193d} regressed the MANO shape and pose parameters from 2D keypoint heatmaps. This was the first end-to-end deep learning based method that can predict both 3D hand shape and pose from RGB images in the wild. 
Zhang~\emph{et al}~\cite{zhang2019end} proposed to use an iterative regression module to regress the MANO parameters in a coarse-to-fine manner.
Baek~\emph{et al}~\cite{baek2019pushing} also exploited iterative refinement. In addition to that, a differentiable renderer was also deployed, which can be supervised by 2D segmentation masks and 3D skeletons.
Hasson~\emph{et al}~\cite{hasson2019learning} exploited the MANO model to solve the task of reconstructing
hands and objects during manipulation. Yang~\emph{et al}~\cite{yang2020bihand} proposed a multi-stage bisected network, which can regress the MANO params using 3D heatmaps and depth map.

{\bf Model-free methods.} In~\cite{moon2020i2l},   Moon~\emph{et al} designed I2L-MeshNet, an image-to-lixel prediction network. 
Many other works are based on graph convolutional network, directly regressing the vertex locations. In~\cite{ge20193d}, Ge~\emph{et al} proposed a graph neural network based method to reconstruct a full 3D mesh of hand surface.  
In~\cite{lim2018simple}, Lim~\emph{et al} proposed an efficient graph convolution, SpiralConv, to process mesh data in the spatial domain.
Leveraging spiral mesh convolutions, Kulon~\emph{et al}~\cite{kulon2020weakly} devised a simple and effective network architecture for monocular 3D hand pose estimation consisting of an image encoder followed by a mesh decoder.
Most recently, Chen~\emph{et al}~\cite{chen2021camera} exploited the similar architecture, with more advanced designs. They divide the camera-space mesh recovery into two sub-tasks, i.e., root-relative mesh recovery and root recovery. To estimate the root-relative mesh, the authors proposed a novel aggregation method to collect effective 2D cues from the image, and then are decoded by a spiral graph convolutional decoder to regress the vertex coordinates.
Apart from graph neural network, Transformer~\cite{vaswani2017attention} has also been introduced into the field of computer vison, solving different tasks~\cite{liu2021swin, carion2020end, yan2022after}. Several methods~\cite{lin2021end, lin2021mesh, hampali2022keypoint, park2022handoccnet} have been proposed for hand pose and mesh reconstruction.

{\bf Hand model personalization.} Tan \emph{et al}~\cite{tan2016fits} and Tkach \emph{et al}~\cite{tkach2017online} studied hand model personalization in the scenario where multiple \emph{depth} images are available, and successfully demonstrated its importance in hand tracking. Hampali~\emph{et al}~\cite{hampali2020honnotate} used the same method to generate annotations when creating a new dataset. However, hand model personlization from RGB images has been underexplored. 
Qian~\emph{et al}~\cite{qian2020html} focused on hand \emph{texture} personalization from RGB images.
While hand model (mesh) personalization is also performed, the effectiveness of mesh personalization is not validated by quantitative results. There is also no investigation on whether the personalized mesh model can be used to improve hand pose estimation.
Moon~\emph{et al}~\cite{moon2020deephandmesh} proposed to personalize each subject using a randomly generated Gaussian vector. The subject ID vectors were generated prior to training and experiments were performed where all subjects in the test set were already seen in the training set. The trained model is only applicable to known subjects and there exists no principle way to handle unseen subjects during the testing phase.
MEgATrack~\cite{han2020megatrack} is a multi-view monochrome egocentric hand tracking system that calibrates the hand model for unseen users, but the calibration is limited to a single hand scaling factor.

To our best knowledge, our work is the \emph{first} to systematically investigate the hand model personalization from RGB images and its benefits to 3D hand pose estimation and mesh reconstruction.

\section{Method}

We first review the MANO hand model which is used extensively in this work, and then propose our identity-aware hand mesh estimation method that takes as input the identity information represented by the hand MANO shape parameters along with the input image. Next, by a slight modification of our method, we propose the baseline which would be compared with.
Lastly, to address the practical use case where the hand model is not provided for the test subject, we propose a novel personalization pipeline that estimates the hand model for an \emph{unseen} subject using only a few unannotated images.

\subsection{MANO Model}
MANO~\cite{romero2017embodied} is a popular parameterized hand model extended from the 3D human model SMPL~\cite{loper2015smpl}.
The MANO model factorizes the hand mesh into two groups of parameters: the shape parameters and the pose parameters. The shape parameters control the intrinsic shape of the hand, e.g., size of the hand, thickness of the fingers, length of the bones, etc. The pose parameters represent the hand pose, i.e., how the hand joints are transformed, which subsequently deforms the hand mesh. Mathematically, the model is defined as below:
\begin{align}
     {\cal M}(\beta, \theta) &= W(T_P(\beta, \theta), J(\beta), \theta, {\cal W})
\label{eq:mano}
\end{align}
where a skinning function $W$ is applied to an articulated mesh with shape $T_P$, joint locations $J$, pose parameter $\theta$, shape parameter $\beta$, and blend weights $\cal W$~\cite{romero2017embodied}.

\subsection{Identity-aware Hand Mesh Estimation}

\begin{figure}[t]
\begin{center}
   \includegraphics[width=0.9\linewidth]{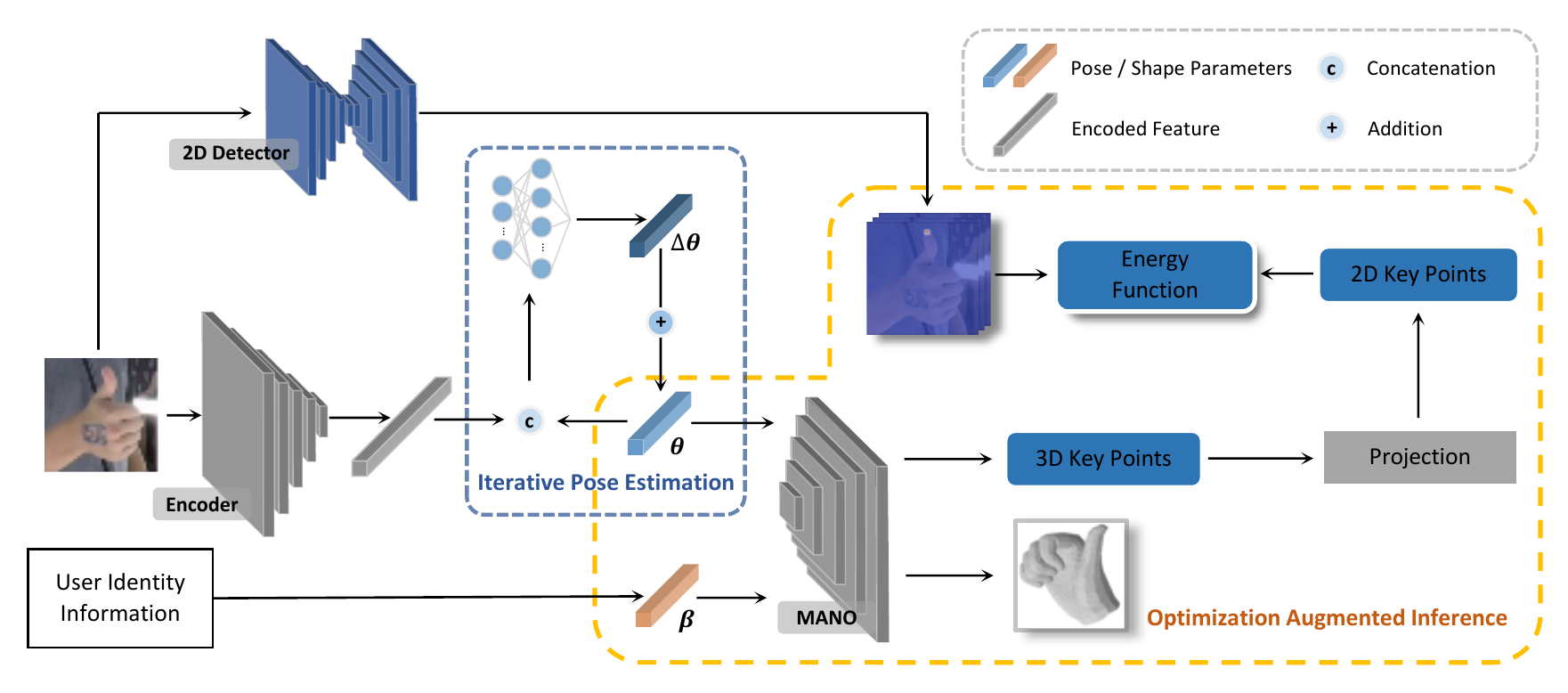}
\end{center}
   \caption{Overview of our proposed identity-aware hand mesh estimation model. The model mainly contains three parts, i.e., the iterative pose regressor, the 2D detector and the optimization module. Note that in our proposed model, along with the RGB image, we also feed the user's identity information, i.e. the ground truth or calibrated MANO shape parameters of the user. }
\label{fig:our_model_overview}
\end{figure}

Existing methods assume that the subject in every image frame is anonymous, even though the input is recorded in a continuous session. To fully leverage the fact that the subject is often fixed within each recording session in real applications, we propose a new hand mesh estimation pipeline. In addition to the input image, we also feed the user's identity information into the network.

There are various ways to represent the identity of a subject. 
The most straightforward method is to label each subject with a unique identifier, such as a high-dimensional random vector~\cite{moon2020deephandmesh}.
However, identity information that does not have physical meaning can be hard for the model to utilize. 
More importantly, models trained with this type of identity information usually only generalize to {\it known} subjects included in the training set. In this work, we are interested in an identity representation that allows generalizing to {\it unseen} subjects.


Inspired by MANO model~\cite{romero2017embodied}, we utilize the MANO shape parameters as the identity information for a specific subject. 
As shown in Fig.~\ref{fig:our_model_overview}, our proposed identity-aware hand mesh estimator 
takes in directly the ground truth or calibrated MANO shape parameters, enabling the network to be subject-aware. 
The main parts of our proposed model is explained as follows.

{\bf{MANO pose parameter regressor.}} Motivated by~\cite{zhang2019end}, the pose parameter $\theta$ is obtained by using an iterative pose regressor. We include the global rotation in  $\theta$, and use the 6D rotation representation~\cite{zhou2019continuity} to  represent the rotation of each joint. With 15 hand joints and the global rotation, $\theta$ is a vector in $\mathbb{R}^{96}$. 
Let ${\cal F}\in \mathbb{R}^N$ denote the image feature after the encoder and $\theta^{(i)}$ denote the estimated pose after $i$ iterations.
Initially, we set $\theta^{(0)}$ as the rotation 6D representation of identity matrices. Then, the pose is predicted iteratively as follows
\begin{align}
      \Delta \theta^{(i)} &= MLP_{\theta} \left({\textrm {cat}}({\cal F}, \theta^{(i-1)})\right) \\
    \theta^{(i)} &=  \Delta \theta^{(i)} \oplus \theta^{(i-1)},
\end{align}
where $\oplus$ means adding the new rotation increment onto the predicted rotation from the previous iteration. The operator $\oplus$ is implemented by transforming both  $\Delta \theta^{(i)}$ and  $\theta^{(i-1)}$ from rotation 6D representations to rotation matrices, then multiplying them, and finally converting the result back to rotation 6D representation. We adopt three iterations in the experiments.


{\bf{Optimization Augmented Inference}.}
During inference time, we can further improve the estimated hand mesh by enforcing the consistency between the 3D pose and the 2D pose predictions. The 2D predictions are obtained via a stacked hourglass-style neural network~\cite{newell2016stacked}.

Let ${\bf x}^{\textrm d}\in {\mathbb R}^{21\times 2}$ denote the 2D keypoints predictions, $f_{\textrm {MANO} }(\cdot)$ represent the mapping function from $(\beta, \theta)$ to 3D keypoints positions, $\mathcal{P}(\cdot)$ denote the projection operator from 3D space to image space, and $\bf r\in{\mathbb R}^{3}$ denote the root-position of the hand. We aim to optimize the following energy function
\begin{equation}
    \mathcal{E}(\theta, {\bf r}, \beta) = \left\Vert {\bf x}^{\textrm d} - {\mathcal{P}}(f_{\textrm {MANO}}(\beta, \theta) +{\bf r}) 
    \right\Vert _2 .
\label{eq:inference_optimization}
\end{equation}
We adopt a two-stage optimization procedure. In the first stage, we optimize $\bf r$ only. In the second stage, we optimize $\theta$ and $\bf r$ jointly. Note that the MANO parameters are not optimized from scratch. The prediction from the MANO parameter regressor is used as the initial guess.

\subsection{Baseline Method}
To further validate that the accuracy improvement of hand mesh reconstruction is a result of leveraging the identity information,
we construct a baseline by slightly modifying our identity-aware model. Instead of feeding ground truth/calibrated shape parameters into the model, we use an extra MLP to regress the shape parameters from the input image. For a fair comparison, all the other modules from our identity-aware model are kept the same in this baseline model. Formally, let ${\cal F}\in \mathbb{R}^N$ denote the image feature produced by the encoder. The MANO shape parameter $\beta \in \mathbb{R}^{10}$ is directly regressed by a multilayer perceptron from ${\cal F}$ as
\begin{equation}
    \beta = MLP_{\beta} ({\cal F}).
\end{equation}


\begin{figure}[hbt!]
\begin{center}
   \includegraphics[width=0.9\linewidth]{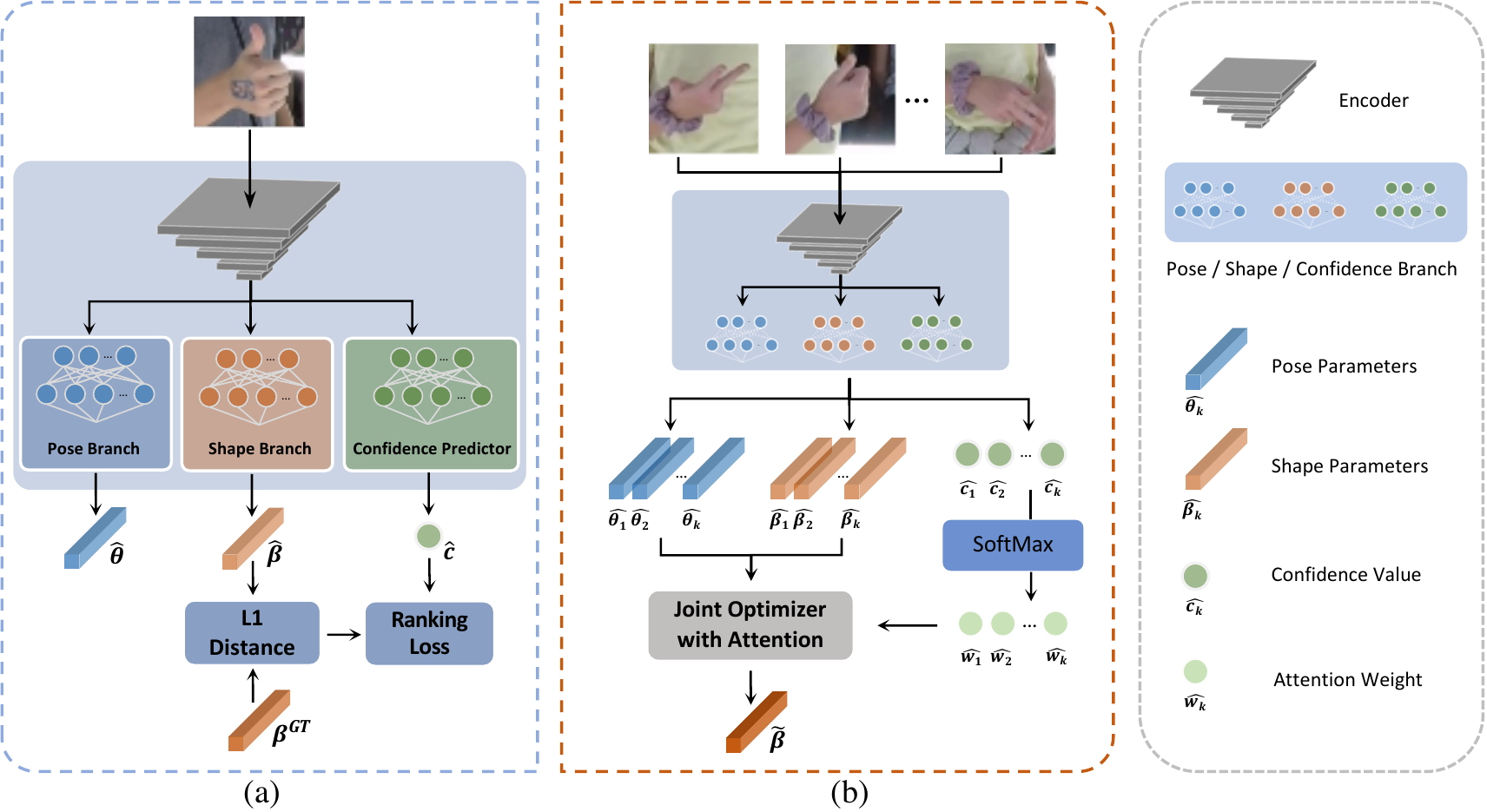}
\end{center}
   \caption{Proposed personalization pipeline with attention mechanism. Images used for personalization capture the same subject who is never seen during training.}
\label{fig:conf_branch}
\end{figure}

\subsection{Personalization Pipeline}
In most practical applications, the test subject is usually \emph{unknown} and there is no corresponding hand model (shape parameters) available for the proposed identity-aware hand mesh estimation pipeline. To handle this practical issue, we propose a novel hand model personalization method, which could calibrate the hand model from a few \emph{unannotated} RGB images.

\subsubsection{Confidence Predictor.} Our personalization pipeline takes in multiple images of a same subject and perform a joint attention-based optimization to get the personalized shape parameter.
Naively, the images can be treated equally and contribute the same weight during the optimization. 
However, images usually differ from each other in terms of quality, view angles, occlusions and so on. 
Thus, the images should be attended with different importance. 
To achieve this goal, we propose a light weight confidence predictor on top of the baseline network, as shown in Fig.~\ref{fig:conf_branch} (a).
The confidence predictor takes as input the feature extracted by the ResNet50 encoder and outputs a scalar via one fully connected layer.
The predicted confidence value indicates the quality of the predicted shape parameter from the input image. Note that our confidence predictor is only trained on the training split. 
Subjects in the test split are different from the training split and are \textbf{not seen} during the training phase.

\subsubsection{Joint Optimization with Attention.} Fig.~\ref{fig:conf_branch} (b) illustrates the whole process during the personalization phase. Denote the collection of $K$ unannotated images from the same user as ${\cal I} = \{I_1, I_2, \cdots, I_K \}$. 
The images are fed into the baseline model equipped with confidence predictor, which outputs $\{c_i, \hat \beta_i, \hat \theta_i\}$ for each image $I_i$, where ${c_i}\in {\mathbb R}$ is the confidence value, $\hat \beta_i, \hat \theta_i$ are the predicted MANO shape and pose parameters. The confidence values $\{c_i\}_{i=1}^K$ then go through a SoftMax layer, which generates the attention weights $\{w_i\}_{i=1}^K$ as following
\begin{equation}
    w_i = \frac{e^{c_i/T}}{\sum_{k=1}^K e^{c_k/T}},
\label{eq:softmax}
\end{equation}
where $T$ is the temperature parameter. Afterwards, $\{w_i, \hat \beta_i, \hat \theta_i \}_{i=1}^K$ are sent into the attention based optimization module, where the following optimization is solved
\begin{equation}
    \displaystyle{\min_{\Tilde  \beta} \sum_{k=1}^K w_k \cdot \lVert {\cal M}(\Tilde \beta, \hat \theta_k) - {\cal M}(\hat \beta_k, \hat \theta_k)  \rVert _F},
\label{eq:weighted_calibration}
\end{equation}
where $\cal M(\cdot)$ is the MANO model. Note that, now all the $K$ images from the same subject share same shape parameter ${\Tilde  \beta}$. After the personalization process, ${\Tilde  \beta}$ would be used as the identity information for the subject.

\subsection{Loss Functions}
\paragraph{The baseline.}
To train the baseline, we apply loss terms on the predicted 3D hand mesh,  following~\cite{chen2021camera}, and also on the predicted MANO shape and pose parameters.

a) Loss functions on hand mesh.
Denote the vertices and faces of the hand mesh as ${\cal V}$ and ${ \Omega}$.
We impose $L1$ loss on the predicted hand mesh, and also deploy  edge length loss and normal loss, following~\cite{chen2021camera}.
The loss functions on the mesh can be expressed as
\begin{equation}
\begin{aligned}
    L_{\rm mesh} &= \sum_{i=1}^{N} \| {\hat {\cal V}_i} - {\cal V}_i  \|_1 \\
    L_{\rm norm} &= \sum_{\omega \in {\Omega}} \sum_{(i,j) \subset \omega}  \left| \frac{{\hat {\cal V}}_i - {\hat {\cal V}}_j}{\|{\hat {\cal V}}_i - {\hat {\cal V}}_j\|_2} \cdot {\bf n}_{\omega} \right| \\
    L_{\rm edge} &= \sum_{\omega \in {\Omega}} \sum_{(i,j) \subset \omega}  \left| \|{\hat {\cal V}}_i - {\hat {\cal V}}_j\|_2 - \|{ {\cal V}}_i - { {\cal V}}_j\|_2 \right|,
\end{aligned}
\end{equation}
where the ${\bf n}_{\omega}$ is the unit normal vector of face $\omega \in \Omega$.

b) Loss function on MANO parameters.
We use $L_{\rm pose} = \|{\hat \theta} - \theta  \|_1$ and 
    $L_{\rm shape} = \|{\hat \beta} - \beta  \|_1$,
where $\theta$ and $\beta$ are ground truth MANO pose and shape parameters. The $\hat \theta$ is the predicted pose parameter from the last iteration of the iterative pose regressor.

c) Loss function on 2D heatmap. A binary cross entropy function is imposed on 2D heatmaps of hand keypoints as in
$
    L_{\rm pose_{2D}} = {\rm BCE}(\hat U, U),
$
where $\hat U$ and $U$ are the predicted and ground truth 2D heatmaps of each keypoint, respectively. The groud-truth heatmap $U$ is generated with a Gaussian distribution. 

The 2D detector is trained by using $L_{\rm pose_{2D}}$. The other parts are trained under the following loss function
\begin{equation}
    L_{\rm total} = L_{\rm mesh} + 0.1 \cdot L_{\rm norm} + L_{\rm edge} + L_{\rm pose} +L_{\rm shape}. 
\label{eq:loss_fun_baseline}
\end{equation}

\paragraph{Our identity-aware model.} Since for our identity-aware model, the subject identity information (the MANO shape parameter) is provided, either ground truth or calibrated, the loss function is given by Eq.~(\ref{eq:loss_fun_ours}) with the shape loss removed,
\begin{equation}
    L_{\rm total}^{'} = L_{\rm mesh} + 0.1 \cdot L_{\rm norm} + L_{\rm edge} + L_{\rm pose}. 
\label{eq:loss_fun_ours}
\end{equation}

\paragraph{Confidence Predictor.} We use margin ranking loss for training of the confidence predictor. Given $N_b$ images in the batch, the baseline model equipped with confidence predictor would output confidence values $\{c_i\}_{i=1}^{N_b}$ and MANO shape parameter predictions $\{\hat \beta_i\}_{i=1}^{N_b}$. With ground truth shape parameters $\{ \beta_i\}_{i=1}^{N_b}$, the difference $l_i$ between the predicted and ground truth shape parameters can be calculated as $l_i = |\beta_i - \hat \beta_i|_1$. We generate $N_b\times (N_b-1)/2$ pairs of $\{(c_i, l_i), (c_j, l_j) \}$, and calculate ranking loss on each pair~\cite{rankingloss}. The total loss is the sum of ranking losses from all pairs.


\section{Experiments}
\subsection{Experimental Setups}
{\bf Datasets.} We conduct experiments on two large-scale public hand pose datasets, i.e., HUMBI~\cite{yu2020humbi} and DexYCB~\cite{chao2021dexycb}. 
There are two major reasons why these two datasets are chosen.
First, they both have a diverse collection of subjects, which allows us to split the datasets into different subject groups for training and evaluating our identity-aware pipeline. More importantly, they annotate the shape parameters of the same subject in a consistent way. Each hand image in the dataset is associated with a subject ID and all the hands from the same subject share the same MANO shape annotation. Note that our method cannot be directly evaluated on other popular benchmarks such as FreiHAND~\cite{zimmermann2019freihand} or InterHand~\cite{Moon_2020_ECCV_InterHand2.6M} because they either do not associate images with subject IDs or guarantee consistent shape parameters for the same subject.

{\bf HUMBI} is a large multiview image dastaset of human body expressions with natural clothing. 
For each hand image, the 3D mesh annotation is provided, along with the fitted MANO parameters. The shape parameters are fitted across all instances of the same subject. This means that the same shape parameters are \emph{shared} among all the hand meshes from the same subject. In our experiments, we use all the right hand images from the released dataset.
We split the dataset into training (90\%) and test (10\%), by subjects. The split results into 269 subjects (474,472 images) in the training set and 30 subjects (50,894 images) in the test set. 
Note that none of the subjects in the test set appear in the training set.

{\bf DexYCB} is a large dataset capturing hand grasping of objects. 
The dataset consists of 582K RGB-D frames over 1,000 sequences of 10 subjects from 8 views. It also provides MANO parameters for each hand image. Same as the HUMBI dataset, the hand shape parameters for each subject are calibrated and fixed throughout each subject's sequences. While object pose estimation is beyond the scope of this work, extra occlusions introduced by the objects makes the DexYCB dataset more challenging for hand mesh estimation. In our experiments, similar to the set up for the HUMBI dataset, we use the provided split in~\cite{chao2021dexycb} which splits the dataset by subjects. In this set up, there are 7, 1, 2 subjects in the training, validation and test set, respectively. \newline

{\bf Metrics for 3D Hand Estimation.} Following the protocol used by existing methods, we use the following two metrics, both in millimeter.

a) \textit { Mean Per Joint Position Error} (MPJPE) measures the Euclidean distance between the root-relative prediction and ground truth 3D hand keypoints.

b) \textit {Mean Per Vertex Position Error} (MPVPE) measures the Euclidean distance between the root-relative prediction and groud-truth 3D hand mesh.

{\bf Metrics for Hand Shape Calibration. } We propose three metrics to evaluate the performance of the calibrated hand shape.

a) \textit{$MSE_{mano}$} measures the mean square error between the estimated MANO shape parameters and the ground truth values.

b) \textit{W-error} measures the mean hand width error between the calibrated hands and the ground truth hands at the flat pose, which is defined as the distance between the metacarpophalangeal joints of index finger and ring finger.

b) \textit{L-error} measures the mean hand length error between the calibrated hands and the ground truth hands at the flat pose, which is defined as the distance between the wrist joint and the tip of middle finger as the hand length.

{\bf Implementation Details.}
We implement our model in PyTorch~\cite{paszke2019pytorch} and deploy ResNet50~\cite{he2016deep} as our encoder. Input images are resized to 224$\times$224 before being fed into the network. We use the Adam optimizer~\cite{kingma2014adam} and a batch size of 32 to train all the models except for the confidence predictor.
For a fair comparison, both the baseline model and our proposed identity-aware model are trained using the same learning rate schedule. On the HUMBI dataset, both models are trained for 15 epochs, with an initial learning rate of 1e-4 which is dropped by a factor of 10 at the 10-th epoch.
On the DexYCB dataset, models are also trained for 15 epochs, with the same initial learning rate, while the learning rate is dropped at the 5-th and 10-th epochs.
With the baseline model trained and frozen, the lightweight confidence predictor is trained with a batch size of 128, with the intuition that larger batch size allows more image pairs to train the ranking loss. 
The temperature parameter is set to 0.33 in Eq.~(\ref{eq:softmax}).
During all the training, input images are augmented with random color jitter and normalization.
In the inference stage, we use the Adam optimizer in PyTorch to optimize Eq.~(\ref{eq:inference_optimization}). 
Specifically, 200 and 60 iterations are performed with learning rate of 1e-2 and 1e-3 in the first and second optimization stages, respectively. On one Titan RTX graphics card, it takes 8 minutes to process all test images (50k) in HUMBI dataset, and 7.5 minutes for those (48k) in DexYCB dataset. We emphasize again that all our experiments are conducted in the scenarios where there is {\bf no overlap} between the subjects in the test set and the training set.

\begin{table}[t]
    \begin{minipage}{1.0\linewidth}
    \caption{Numerical results on DexYCB and HUMBI datasets.}
    \centering
    \begin{tabular}{ l c c c c} 
     \hline 
    \toprule
    \multirow{2}{*}{Method}               & \multicolumn{2}{c}{ DexYCB } & \multicolumn{2}{c}{ HUMBI } \\
    \cmidrule(lr){2-3} \cmidrule(lr){4-5} 
    {}                     & MPJPE $\downarrow$    & MPVPE $\downarrow$ & \;\;\;\;MPJPE $\downarrow$    & MPVPE $\downarrow$\\
    \hline
    {CMR-PG~\cite{chen2021camera}} & 20.34 & 19.88 &\;\;\;\;11.64&11.37\\
    \midrule
    \multicolumn{5}{c} {Without Optimization at Inference Time} \\
    \hline
    Baseline               & 21.58                 & 20.95     &\;\;\;\;12.13  &11.82\\
    Ours, GT shape     & 18.83                 & 18.27     &\;\;\;\;11.41  &11.11\\
    Ours, Calibrated     & 18.97                 & 18.42  &\;\;\;\;11.51 &11.21\\
    \hline
    \multicolumn{5}{c} {With Optimization at Inference Time} \\
    \hline
    Baseline               & 18.03                 & 17.92 &\;\;\;\;10.75&10.60\\
    Ours, GT shape  & 16.60                 & 16.29 &\;\;\;\;10.17&9.94\\
    Ours, Calibrated       & \bf{16.81}                & \bf{16.55} &\;\;\;\;\bf{10.31}&\bf{10.28}\\
    \bottomrule
    \label{table:results_on_both_datasets}
    \end{tabular}      
    \end{minipage}
\end{table}

\begin{table}[t]
    \begin{minipage}{.6\linewidth}
      \caption{Comparison with existing methods on Dex-YCB. }
      \centering
    \begin{tabular}{c c c}
        \toprule
         Methods & MPJPE$\downarrow$ & MPVPE $\downarrow$  \\
        \hline
        Boukhayma et al.~\cite{boukhayma20193d}& 27.94 &27.28 \\
        {Spurr~\emph{et al}~\cite{spurr2020weakly} + ResNet50} & 22.71 & - \\
        {Spurr~\emph{et al}~\cite{spurr2020weakly} + HRNet32}  &22.26 & - \\
        {Boukhayma et al.~\cite{boukhayma20193d} \textsuperscript{\textdagger}} & 21.20 & 21.56\\
        CMR-PG~\cite{chen2021camera} & 20.34 & 19.88 \\
        Metro~\cite{lin2021end} & 19.05 & 17.71 \\
        
        \hline
        Ours, Calibrated &  {\bf 16.81}    &{\bf 16.55} \\
        \bottomrule
    \label{table:additional_results_on_dex_ycb}
    \end{tabular}
    \end{minipage}
    \begin{minipage}{.4\linewidth}
      \centering
        \caption{Performance of hand model calibration.}
        \begin{tabular}{ c c c  } 
        \toprule
        Metrics     & HUMBI\;  &\;DexYCB\;\\
        \hline
        MSE\text{$_{mano}$}  &  0.07      &0.04  \\
        W-error (mm)     &  0.88      & 1.02 \\
        L-error (mm)    &   1.71     & 1.20 \\
        \bottomrule
        \label{table:hand_shape}
        \end{tabular}
    \end{minipage} 
\end{table}


\subsection{Quantitative Evaluation}



{ \bf 3D Hand Estimation.} 
We evaluate the benefit of our pipeline under two settings, i.e., with and without the optimization module during inference time. 
As shown in Table~\ref{table:results_on_both_datasets}, our proposed pipeline improves the baseline consistently across different datasets.
With calibrated hand model, our proposed method can achieve close performance to that with ground truth hand model, which validates the effectiveness of our personalization pipeline. Furthermore, our method also achieves the state-of-the-art performance, as shown by Table~\ref{table:results_on_both_datasets} and Table~\ref{table:additional_results_on_dex_ycb}. To ensure fair comparison, same data augmentation are applied to all the methods, i.e.,  random color jitter and
normalization. All the models are trained for 15 epochs including ours, with the exception of Metro~\cite{lin2021end} which is trained for 70 epochs, as transformers are much harder to converge. The superscript \textsuperscript{\textdagger} in Table~\ref{table:additional_results_on_dex_ycb} means adding our optimization module on top of the original method. It shows that our optimization module can be generalized to other model-based methods efficiently. We emphasize that,
\emph{none} of the existing methods produces consistent shape estimation across images originating from the same subject. In contrast, our method guarantees shape consistency with zero hand shape variation.

{ \bf Hand Shape Calibration.} Table~\ref{table:hand_shape} reports the performance of our personalization pipeline, which achieves less than 2 mm in terms of hand width and hand length errors, by calibrating on 20 unannotated images. 

\begin{figure}[t]
\begin{center}
   \includegraphics[width=0.95\linewidth]{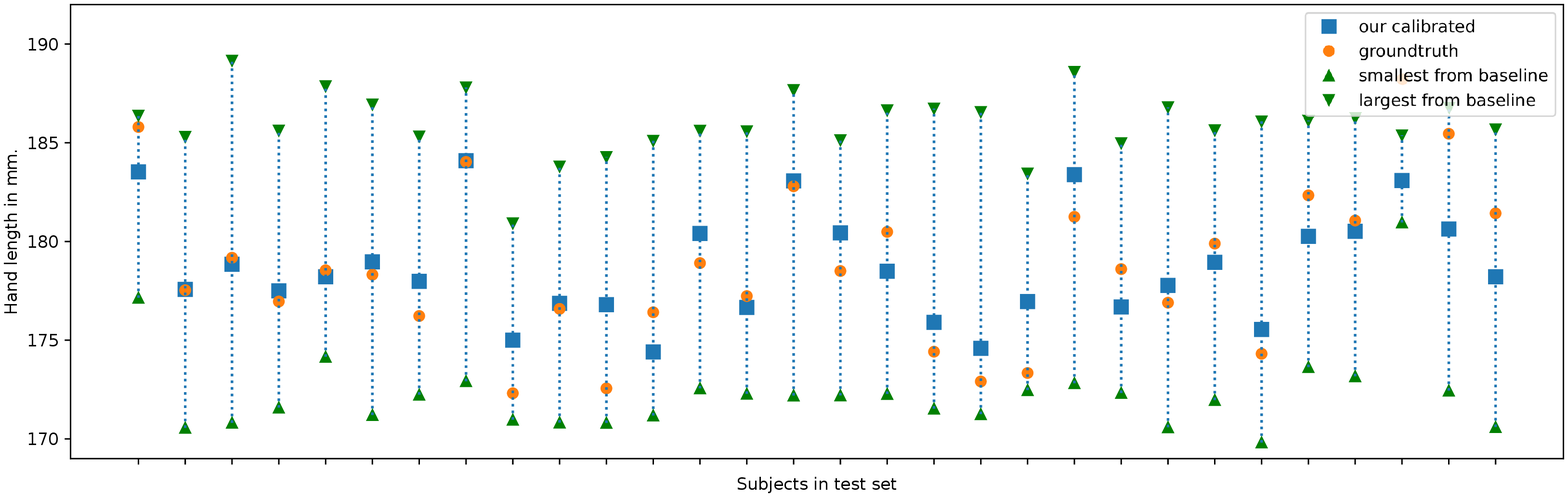}
\end{center}
   \caption{Hand shape consistency comparison between our proposed method and the baseline. The x-axis corresponds to different subjects in the test dataset, while the y-axis corresponds to the length of the hand of each subject.}
\label{fig:shape_consistency_comp}

\begin{center}
   \includegraphics[width=0.95\linewidth]{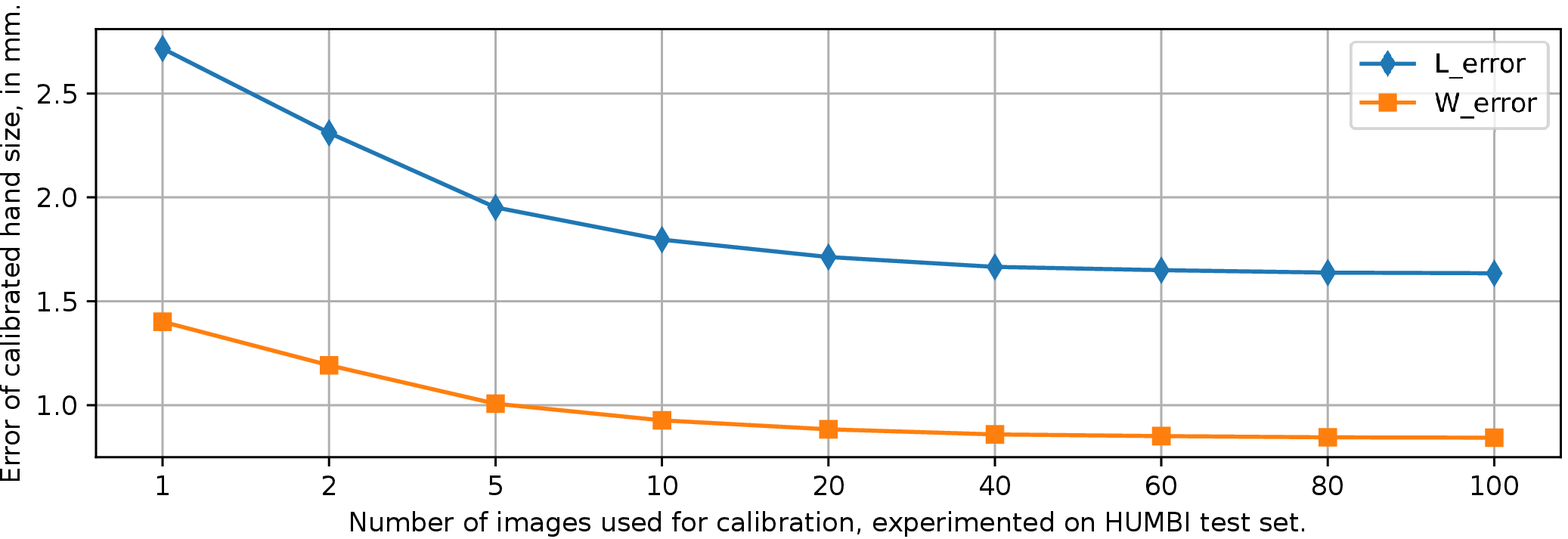}
\end{center}
\caption{Impact of the number of images used in calibration.}
\label{fig:numer_of_images}
\end{figure}

Our proposed method inherently guarantees the hand shape consistency among different images from the same subject. Fig.~\ref{fig:shape_consistency_comp} demonstrates this advantage of our method over the baseline model.
As shown in Fig.~\ref{fig:shape_consistency_comp}, for a specific subject, the baseline model outputs hand meshes with big variations in terms of hand length, up to 20 mm.
This is because the baseline model is subject-agnostic and predicts hand shape parameters based on a single input image. 
Even if the input images are from the same user, the baseline model could predict hand meshes with big variations in size.
In contrast, our proposed method outputs consistent hand shape inherently, with \emph{zero} hand shape variation across images from the same user. Also shown in Fig.~\ref{fig:shape_consistency_comp}, the hand size calibrated by our proposed method stays close to the ground truth hand size in most cases.


\subsection{Ablation Study}

{\bf Number of images used for personalization.} Fig.~\ref{fig:numer_of_images} shows hand size errors (in mm) when different number of images are utilized during calibration. With $K=20$ images, the hand model can already be well calibrated with length error less than 2 mm and width error  less than 1 mm. In all the other experiments, we use $K=20$ images for hand model calibration.

\begin{table}[t]
\caption{Effectiveness of confidence-valued based attention mechanism.}
\begin{center}
\begin{tabular}{ c ccc } 
\toprule
Metrics     & \;\;MSE{$_{mano}$} \;\;  &\;\;W-error (mm) \;\;&\;\; L-error (mm)\;\; \\
\hline
No attention & 0.084 &1.00 &1.93 \\
Ours, with attention &0.070&0.88&1.71\\
\hline
Improvement & 16\%&12\% &11\%\\
\bottomrule
\label{table:hand_shape_different_calibartion_method}
\end{tabular}

\caption{Evaluating models trained with 3D keypoints instead of mesh supervision on DexYCB and HUMBI datasets.}
\begin{tabular}{ l c c c c} 
 \toprule
\multirow{2}{*}{Method}               & \multicolumn{2}{c}{ DexYCB } & \multicolumn{2}{c}{ HUMBI } \\
\cmidrule(lr){2-3} \cmidrule(lr){4-5} 
{}                     &\;\; MPJPE$\downarrow \;\;$    & \;\;MPVPE $\downarrow$ \;\;&  \;\;\;\;\;\;MPJPE $\downarrow$    \;\;&\;\; MPVPE $\downarrow$\;\;\\
\midrule
\multicolumn{5}{c} {Without Optimization at Inference Time} \\
\hline
Baseline               & 21.85                 & 20.26     &\;\;\;\;12.34  &12.02\\
Ours, GT Shape \;\;   & 18.92                 & 18.35     &\;\;\;\;11.61  &11.30\\
\hline
\multicolumn{5}{c} {With Optimization at Inference Time} \\
\hline
Baseline               & 17.71                 & 17.58 &\;\;\;\;10.80&10.95\\
Ours, GT Shape  & {\bf 16.63}                 & {\bf 16.32} &\;\;\;\;{\bf 10.37}& {\bf 10.12}\\
\bottomrule
\label{table:trained_with_keypoints}
\end{tabular}
\end{center}
\end{table}

\noindent {\bf Attention during calibration.} 
During the calibration, different weights are imposed across the input images according to their confidence values, as formulated in Eq.~(\ref{eq:weighted_calibration}). We compare the calibration performance of our attention-based method with the non-attention method, as shown in Table.~\ref{table:hand_shape_different_calibartion_method}. Specifically, non-attention means to treat each image equally and set $w_i = 1/K$ in Eq.~(\ref{eq:weighted_calibration}) for all images.
As shown by Table.~\ref{table:hand_shape_different_calibartion_method}, our attention-based calibration can improve the performance by a noticeable margin comparing to the naive calibration.

\noindent {\bf Optimization augmented inference from scratch.} In this experiment, we remove the MANO parameter regressor from the model in Fig.~\ref{fig:our_model_overview}. Without being initialized by the MANO parameter regressor, the initial pose is set to the neutral pose prior to optimization. This pure optimization procedure results in an MJPJE $>50$mm on both DexYCB and HUMBI datasets. This validates the necessity of the MANO parameter regressor, which can give good initial values of MANO parameters for later optimization. 

\noindent {\bf Training model with 3D keypoints instead of 3D mesh supervision.} In Table.~\ref{table:trained_with_keypoints}, we report the performance of the baseline and our proposed identity-aware model when trained with 3D keypoints supervision, instead of 3D mesh. Under this setting, our identity-aware  method still improves the accuracy.

\begin{figure}[t]

\begin{center}
  \includegraphics[width=0.95\linewidth]{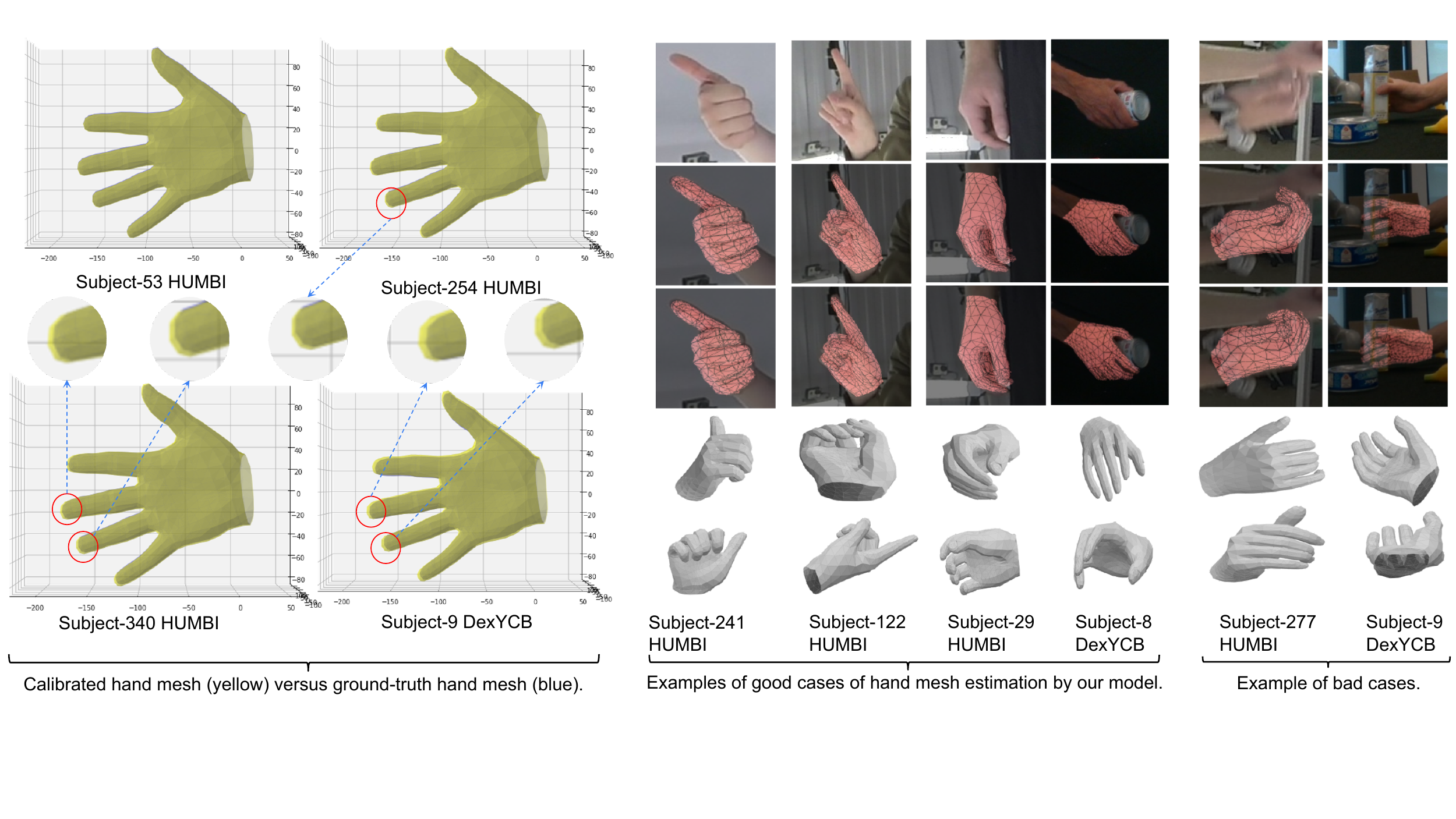}
\end{center}
\vspace{-0.3in}
\caption{Qualitative results. a) Left: calibrated hand model versus ground truth hand model. b) Right: visualization of our identity-aware hand mesh estimator. From top row to bottom row are the input RGB images, the projected ground truth meshes, the projected predicted meshes, and the predicted meshes viewed from two different angles.}
\label{fig:qualitative_results}
\end{figure}

 {\bf Qualitative Results.} The qualitative results of our personalization method and the identity-aware hand mesh estimator are shown in Fig.~\ref{fig:qualitative_results}. On the left side, it can be seen that the calibrated hand mesh is very close to the ground truth hand mesh.
On the right side, qualitative results of our identity-aware model are demonstrated. When generating the third row,  we align the predicted mesh with ground truth root position before projecting the mesh back to the image space. As seen from Fig.~\ref{fig:qualitative_results}, our model can robustly recover the hand mesh under moderate occlusion and can handle a wide range of hand poses. 

{\bf Limitations.}
The guarantee of consistent hand shape primarily comes from explicitly incorporating a 3D hand model i.e.,  the MANO in our pipeline. A
future direction is to explore model free approaches to enforce shape consistency at inference time. We also observe
that images with severe occlusions and blurs may affect the
quality of shape calibration. We currently mitigate this issue by predicting confidence values, which helps lower the
importance of these suboptimal images greatly. A better approach might be to detect and remove these images prior to
the calibration step.

\section{Conclusion}
In this paper, we propose an \emph{identity-aware} hand mesh estimation pipeline for 3D hand mesh recovery from monocular images.
Different from existing methods which estimate the hand mesh anonymously, our method leverages the fact that the user is usually unchanged in real applications and identity information of the subject can be utilized for 3D hand mesh recovery.
More specifically, our model not only takes as input the RGB image, but also the identity information represented by the intrinsic shape parameters of the subject. 
We also design a novel personalization pipeline, through which the intrinsic shape parameters of an \emph{unknown} subject can be calibrated from a few RGB images.
With the personalization pipeline, our model can operate in scenarios where ground truth hand shape parameters of subjects are not provided, which are common in real world AR/VR applications.
We experimented on two large-scale public datasets, HUMBI and DexYCB, demonstrating the state-of-the-art performance of our proposed method.

\bibliographystyle{splncs04}
\bibliography{eg}

\begin{thebibliography}{10}
\providecommand{\url}[1]{\texttt{#1}}
\providecommand{\urlprefix}{URL }
\providecommand{\doi}[1]{https://doi.org/#1}

\bibitem{athitsos2003estimating}
Athitsos, V., Sclaroff, S.: Estimating 3d hand pose from a cluttered image. In:
  2003 IEEE Computer Society Conference on Computer Vision and Pattern
  Recognition, 2003. Proceedings. vol.~2, pp. II--432. IEEE (2003)

\bibitem{baek2019pushing}
Baek, S., Kim, K.I., Kim, T.K.: Pushing the envelope for rgb-based dense 3d
  hand pose estimation via neural rendering. In: Proceedings of the IEEE/CVF
  Conference on Computer Vision and Pattern Recognition. pp. 1067--1076 (2019)

\bibitem{beddiar2020vision}
Beddiar, D.R., Nini, B., Sabokrou, M., Hadid, A.: Vision-based human activity
  recognition: a survey. Multimedia Tools and Applications  \textbf{79}(41),
  30509--30555 (2020)

\bibitem{boukhayma20193d}
Boukhayma, A., Bem, R.d., Torr, P.H.: 3d hand shape and pose from images in the
  wild. In: Proceedings of the IEEE/CVF Conference on Computer Vision and
  Pattern Recognition. pp. 10843--10852 (2019)

\bibitem{cai2018weakly}
Cai, Y., Ge, L., Cai, J., Yuan, J.: Weakly-supervised 3d hand pose estimation
  from monocular rgb images. In: Proceedings of the European Conference on
  Computer Vision (ECCV). pp. 666--682 (2018)

\bibitem{cao2019openpose}
Cao, Z., Hidalgo, G., Simon, T., Wei, S.E., Sheikh, Y.: Openpose: realtime
  multi-person 2d pose estimation using part affinity fields. IEEE transactions
  on pattern analysis and machine intelligence  \textbf{43}(1),  172--186
  (2019)

\bibitem{carion2020end}
Carion, N., Massa, F., Synnaeve, G., Usunier, N., Kirillov, A., Zagoruyko, S.:
  End-to-end object detection with transformers. In: European conference on
  computer vision. pp. 213--229. Springer (2020)

\bibitem{chao2021dexycb}
Chao, Y.W., Yang, W., Xiang, Y., Molchanov, P., Handa, A., Tremblay, J.,
  Narang, Y.S., Van~Wyk, K., Iqbal, U., Birchfield, S., et~al.: Dexycb: A
  benchmark for capturing hand grasping of objects. In: Proceedings of the
  IEEE/CVF Conference on Computer Vision and Pattern Recognition. pp.
  9044--9053 (2021)

\bibitem{chen2021camera}
Chen, X., Liu, Y., Ma, C., Chang, J., Wang, H., Chen, T., Guo, X., Wan, P.,
  Zheng, W.: Camera-space hand mesh recovery via semantic aggregation and
  adaptive 2d-1d registration. In: Proceedings of the IEEE/CVF Conference on
  Computer Vision and Pattern Recognition. pp. 13274--13283 (2021)

\bibitem{chen2020nonparametric}
Chen, Y., Ma, H., Kong, D., Yan, X., Wu, J., Fan, W., Xie, X.: Nonparametric
  structure regularization machine for 2d hand pose estimation. In: Proceedings
  of the IEEE/CVF Winter Conference on Applications of Computer Vision. pp.
  381--390 (2020)

\bibitem{ge20193d}
Ge, L., Ren, Z., Li, Y., Xue, Z., Wang, Y., Cai, J., Yuan, J.: 3d hand shape
  and pose estimation from a single rgb image. In: Proceedings of the IEEE/CVF
  Conference on Computer Vision and Pattern Recognition. pp. 10833--10842
  (2019)

\bibitem{Ge_2018_ECCV}
Ge, L., Ren, Z., Yuan, J.: Point-to-point regression pointnet for 3d hand pose
  estimation. In: Proceedings of the European Conference on Computer Vision
  (ECCV) (September 2018)

\bibitem{hampali2020honnotate}
Hampali, S., Rad, M., Oberweger, M., Lepetit, V.: Honnotate: A method for 3d
  annotation of hand and object poses. In: Proceedings of the IEEE/CVF
  conference on computer vision and pattern recognition. pp. 3196--3206 (2020)

\bibitem{hampali2022keypoint}
Hampali, S., Sarkar, S.D., Rad, M., Lepetit, V.: Keypoint transformer: Solving
  joint identification in challenging hands and object interactions for
  accurate 3d pose estimation. In: Proceedings of the IEEE/CVF Conference on
  Computer Vision and Pattern Recognition. pp. 11090--11100 (2022)

\bibitem{han2020megatrack}
Han, S., Liu, B., Cabezas, R., Twigg, C.D., Zhang, P., Petkau, J., Yu, T.H.,
  Tai, C.J., Akbay, M., Wang, Z., et~al.: Megatrack: monochrome egocentric
  articulated hand-tracking for virtual reality. ACM Transactions on Graphics
  (TOG)  \textbf{39}(4),  87--1 (2020)

\bibitem{hasson2019learning}
Hasson, Y., Varol, G., Tzionas, D., Kalevatykh, I., Black, M.J., Laptev, I.,
  Schmid, C.: Learning joint reconstruction of hands and manipulated objects.
  In: Proceedings of the IEEE/CVF Conference on Computer Vision and Pattern
  Recognition. pp. 11807--11816 (2019)

\bibitem{he2016deep}
He, K., Zhang, X., Ren, S., Sun, J.: Deep residual learning for image
  recognition. In: Proceedings of the IEEE conference on computer vision and
  pattern recognition. pp. 770--778 (2016)

\bibitem{kingma2014adam}
Kingma, D.P., Ba, J.: Adam: A method for stochastic optimization. arXiv
  preprint arXiv:1412.6980  (2014)

\bibitem{kong2019adaptive}
Kong, D., Chen, Y., Ma, H., Yan, X., Xie, X.: Adaptive graphical model network
  for 2d handpose estimation. arXiv preprint arXiv:1909.08205  (2019)

\bibitem{kong2020rotation}
Kong, D., Ma, H., Chen, Y., Xie, X.: Rotation-invariant mixed graphical model
  network for 2d hand pose estimation. In: Proceedings of the IEEE/CVF Winter
  Conference on Applications of Computer Vision. pp. 1546--1555 (2020)

\bibitem{kong2020sia}
Kong, D., Ma, H., Xie, X.: Sia-gcn: A spatial information aware graph neural
  network with 2d convolutions for hand pose estimation. arXiv preprint
  arXiv:2009.12473  (2020)

\bibitem{kulon2020weakly}
Kulon, D., Guler, R.A., Kokkinos, I., Bronstein, M.M., Zafeiriou, S.:
  Weakly-supervised mesh-convolutional hand reconstruction in the wild. In:
  Proceedings of the IEEE/CVF Conference on Computer Vision and Pattern
  Recognition. pp. 4990--5000 (2020)

\bibitem{lim2018simple}
Lim, I., Dielen, A., Campen, M., Kobbelt, L.: A simple approach to intrinsic
  correspondence learning on unstructured 3d meshes. In: Proceedings of the
  European Conference on Computer Vision (ECCV) Workshops. pp.~0--0 (2018)

\bibitem{lin2021end}
Lin, K., Wang, L., Liu, Z.: End-to-end human pose and mesh reconstruction with
  transformers. In: Proceedings of the IEEE/CVF Conference on Computer Vision
  and Pattern Recognition. pp. 1954--1963 (2021)

\bibitem{lin2021mesh}
Lin, K., Wang, L., Liu, Z.: Mesh graphormer. In: Proceedings of the IEEE/CVF
  International Conference on Computer Vision. pp. 12939--12948 (2021)

\bibitem{liu2021swin}
Liu, Z., Lin, Y., Cao, Y., Hu, H., Wei, Y., Zhang, Z., Lin, S., Guo, B.: Swin
  transformer: Hierarchical vision transformer using shifted windows. In:
  Proceedings of the IEEE/CVF International Conference on Computer Vision. pp.
  10012--10022 (2021)

\bibitem{loper2015smpl}
Loper, M., Mahmood, N., Romero, J., Pons-Moll, G., Black, M.J.: Smpl: A skinned
  multi-person linear model. ACM transactions on graphics (TOG)
  \textbf{34}(6),  1--16 (2015)

\bibitem{ma2021transfusion}
Ma, H., Chen, L., Kong, D., Wang, Z., Liu, X., Tang, H., Yan, X., Xie, Y., Lin,
  S.Y., Xie, X.: Transfusion: Cross-view fusion with transformer for 3d human
  pose estimation. arXiv preprint arXiv:2110.09554  (2021)

\bibitem{haoyu22ppt}
Ma, H., Wang, Z., Chen, Y., Kong, D., Chen, L., Liu, X., Yan, X., Tang, H.,
  Xie, X.: Ppt: token-pruned pose transformer for monocular and multi-view
  human pose estimation (2022). \doi{10.48550/ARXIV.2209.08194},
  \url{https://arxiv.org/abs/2209.08194}

\bibitem{moon2018v2v}
Moon, G., Chang, J.Y., Lee, K.M.: V2v-posenet: Voxel-to-voxel prediction
  network for accurate 3d hand and human pose estimation from a single depth
  map. In: Proceedings of the IEEE conference on computer vision and pattern
  Recognition. pp. 5079--5088 (2018)

\bibitem{moon2020i2l}
Moon, G., Lee, K.M.: I2l-meshnet: Image-to-lixel prediction network for
  accurate 3d human pose and mesh estimation from a single rgb image. In:
  Computer Vision--ECCV 2020: 16th European Conference, Glasgow, UK, August
  23--28, 2020, Proceedings, Part VII 16. pp. 752--768. Springer (2020)

\bibitem{moon2020deephandmesh}
Moon, G., Shiratori, T., Lee, K.M.: Deephandmesh: A weakly-supervised deep
  encoder-decoder framework for high-fidelity hand mesh modeling. In: European
  Conference on Computer Vision. pp. 440--455. Springer (2020)

\bibitem{Moon_2020_ECCV_InterHand2.6M}
Moon, G., Yu, S.I., Wen, H., Shiratori, T., Lee, K.M.: Interhand2.6m: A dataset
  and baseline for 3d interacting hand pose estimation from a single rgb image.
  In: European Conference on Computer Vision (ECCV) (2020)

\bibitem{newell2016stacked}
Newell, A., Yang, K., Deng, J.: Stacked hourglass networks for human pose
  estimation. In: European conference on computer vision. pp. 483--499.
  Springer (2016)

\bibitem{park2022handoccnet}
Park, J., Oh, Y., Moon, G., Choi, H., Lee, K.M.: Handoccnet: Occlusion-robust
  3d hand mesh estimation network. In: Proceedings of the IEEE/CVF Conference
  on Computer Vision and Pattern Recognition. pp. 1496--1505 (2022)

\bibitem{paszke2019pytorch}
Paszke, A., Gross, S., Massa, F., Lerer, A., Bradbury, J., Chanan, G., Killeen,
  T., Lin, Z., Gimelshein, N., Antiga, L., et~al.: Pytorch: An imperative
  style, high-performance deep learning library. Advances in neural information
  processing systems  \textbf{32},  8026--8037 (2019)

\bibitem{rankingloss}
Pytorch: Pytorch margin ranking loss (2022),
  \url{https://pytorch.org/docs/stable/generated/torch.nn.MarginRankingLoss.html}

\bibitem{qian2020html}
Qian, N., Wang, J., Mueller, F., Bernard, F., Golyanik, V., Theobalt, C.: Html:
  A parametric hand texture model for 3d hand reconstruction and
  personalization. In: European Conference on Computer Vision. pp. 54--71.
  Springer (2020)

\bibitem{romero2017embodied}
Romero, J., Tzionas, D., Black, M.J.: Embodied hands: Modeling and capturing
  hands and bodies together. ACM Transactions on Graphics (ToG)
  \textbf{36}(6),  1--17 (2017)

\bibitem{simon2017hand}
Simon, T., Joo, H., Matthews, I., Sheikh, Y.: Hand keypoint detection in single
  images using multiview bootstrapping. In: Proceedings of the IEEE conference
  on Computer Vision and Pattern Recognition. pp. 1145--1153 (2017)

\bibitem{simonyan2014very}
Simonyan, K., Zisserman, A.: Very deep convolutional networks for large-scale
  image recognition. arXiv preprint arXiv:1409.1556  (2014)

\bibitem{spurr2020weakly}
Spurr, A., Iqbal, U., Molchanov, P., Hilliges, O., Kautz, J.: Weakly supervised
  3d hand pose estimation via biomechanical constraints. In: Computer
  Vision--ECCV 2020: 16th European Conference, Glasgow, UK, August 23--28,
  2020, Proceedings, Part XVII 16. pp. 211--228. Springer (2020)

\bibitem{tan2016fits}
Tan, D.J., Cashman, T., Taylor, J., Fitzgibbon, A., Tarlow, D., Khamis, S.,
  Izadi, S., Shotton, J.: Fits like a glove: Rapid and reliable hand shape
  personalization. In: Proceedings of the IEEE conference on computer vision
  and pattern recognition. pp. 5610--5619 (2016)

\bibitem{tkach2017online}
Tkach, A., Tagliasacchi, A., Remelli, E., Pauly, M., Fitzgibbon, A.: Online
  generative model personalization for hand tracking. ACM Transactions on
  Graphics (ToG)  \textbf{36}(6),  1--11 (2017)

\bibitem{vaswani2017attention}
Vaswani, A., Shazeer, N., Parmar, N., Uszkoreit, J., Jones, L., Gomez, A.N.,
  Kaiser, {\L}., Polosukhin, I.: Attention is all you need. Advances in neural
  information processing systems  \textbf{30} (2017)

\bibitem{wang2018mask}
Wang, Y., Peng, C., Liu, Y.: Mask-pose cascaded cnn for 2d hand pose estimation
  from single color image. IEEE Transactions on Circuits and Systems for Video
  Technology  \textbf{29}(11),  3258--3268 (2018)

\bibitem{zheGPA}
Wang, Z., Chen, L., Rathore, S., Shin, D., Fowlkes, C.: Geometric pose
  affordance: 3d human pose with scene constraints. In: Arxiv 1905.07718 (2019)

\bibitem{zhecross}
Wang, Z., Shin, D., Fowlkes, C.: Predicting camera viewpoint improves
  cross-dataset generalization for 3d human pose estimation. In: ECCV 3DPW
  workshop (2020)

\bibitem{zhebestofboth}
Wang, Z., Yang, J., Fowlkes, C.: The best of both worlds: Combining model-based
  and nonparametric approaches for 3d human body estimation. In: CVPR ABAW
  workshop (2022)

\bibitem{yan2022after}
Yan, X., Tang, H., Sun, S., Ma, H., Kong, D., Xie, X.: After-unet: Axial fusion
  transformer unet for medical image segmentation. In: Proceedings of the
  IEEE/CVF Winter Conference on Applications of Computer Vision. pp. 3971--3981
  (2022)

\bibitem{yang2020bihand}
Yang, L., Li, J., Xu, W., Diao, Y., Lu, C.: Bihand: Recovering hand mesh with
  multi-stage bisected hourglass networks. arXiv preprint arXiv:2008.05079
  (2020)

\bibitem{yu2020humbi}
Yu, Z., Yoon, J.S., Lee, I.K., Venkatesh, P., Park, J., Yu, J., Park, H.S.:
  Humbi: A large multiview dataset of human body expressions. In: Proceedings
  of the IEEE/CVF Conference on Computer Vision and Pattern Recognition. pp.
  2990--3000 (2020)

\bibitem{zhang2019end}
Zhang, X., Li, Q., Mo, H., Zhang, W., Zheng, W.: End-to-end hand mesh recovery
  from a monocular rgb image. In: Proceedings of the IEEE/CVF International
  Conference on Computer Vision. pp. 2354--2364 (2019)

\bibitem{zhou2019continuity}
Zhou, Y., Barnes, C., Lu, J., Yang, J., Li, H.: On the continuity of rotation
  representations in neural networks. In: Proceedings of the IEEE/CVF
  Conference on Computer Vision and Pattern Recognition. pp. 5745--5753 (2019)

\bibitem{zb2017hand}
Zimmermann, C., Brox, T.: Learning to estimate 3d hand pose from single rgb
  images. Tech. rep., arXiv:1705.01389 (2017),
  \url{https://lmb.informatik.uni-freiburg.de/projects/hand3d/},
  https://arxiv.org/abs/1705.01389

\bibitem{zimmermann2017learning}
Zimmermann, C., Brox, T.: Learning to estimate 3d hand pose from single rgb
  images. In: Proceedings of the IEEE international conference on computer
  vision. pp. 4903--4911 (2017)

\bibitem{zimmermann2019freihand}
Zimmermann, C., Ceylan, D., Yang, J., Russell, B., Argus, M., Brox, T.:
  Freihand: A dataset for markerless capture of hand pose and shape from single
  rgb images. In: Proceedings of the IEEE/CVF International Conference on
  Computer Vision. pp. 813--822 (2019)

\end{thebibliography}

\end{document}